\documentclass[letterpaper, 10 pt, conference]{ieeeconf}  

\IEEEoverridecommandlockouts                              

\usepackage{textcomp}
\usepackage{stfloats}
\usepackage{amsmath,amsfonts}
\usepackage{algorithmic}
\usepackage{algorithm}
\usepackage{array}
\usepackage[caption=false,font=normalsize,labelfont=sf,textfont=sf]{subfig}
\usepackage{verbatim}
\usepackage{graphicx}
\usepackage{hyperref}
\usepackage{cite}
\usepackage{threeparttable}
\usepackage{multirow}
\usepackage{graphicx}
\usepackage{float}
\usepackage{hyperref}
\usepackage{etoolbox}

\title{\LARGE \bf
Implicit Articulated Robot Morphology Modeling with \\Configuration Space Neural Signed Distance Functions
}
\author{Yiting Chen$^{*,4}$\hspace{1cm} Xiao Gao$^{\dagger, 1}$\hspace{1cm} Kunpeng Yao$^{\dagger, 5}$\hspace{1cm} Loïc Niederhauser$^{1}$\\
Yasemin Bekiroglu$^{2,3}$\hspace{1cm} Aude Billard$^{1}$
\thanks{This project was funded by ERC Advanced Grant, Skill Acquisition in Humans and Robots (SAHR) under Grant 741945. K.Y. received funding from the Swiss National Science Foundation (SNSF), project ID: P500PT 217882. Y. Bekiroglu received funding from Chalmers AI Research Center (CHAIR) and Chalmers Gender Initiative for Excellence (Genie).}
\thanks{$^{*}$Work done during the internship at LASA, EPFL$^{1}.$ $^{\dagger}$Equal correspondence. $^{1}$The Learning Algorithms and Systems Laboratory (LASA), \'Ecole Polytechnique F\'ed\'erale de Lausanne (EPFL), Switzerland {\small \tt \{name.surname\}@epfl.ch} $^{2}$ Department of Electrical Engineering, Chalmers University of Technology, Sweden
$^{3}$ Department of Computer Science, University College London, UK $^{4}$ Department of Computer Science, Rice University, USA $^{5}$ Department of Mechanical Engineering, Massachusetts Institute of Technology, USA}%
}
\begin{document}
\maketitle
\begin{abstract}
In this paper, we introduce a novel approach to implicitly encode precise robot morphology using forward kinematics based on a configuration space signed distance function. Our proposed Robot Neural Distance Function (RNDF) optimizes the balance between computational efficiency and accuracy for signed distance queries conditioned on the robot's configuration for each link. 
Compared to the baseline method, the proposed approach achieves an 81.1\%  reduction in distance error while utilizing only 47.6\% of model parameters. Its parallelizable and differentiable nature provides direct access to joint-space derivatives, enabling a seamless connection between robot planning in Cartesian task space and configuration space. These features make RNDF an ideal surrogate model for general robot optimization and learning in 3D spatial planning tasks. Specifically, we apply RNDF to robotic arm-hand modeling and demonstrate its potential as a core platform for whole-arm, collision-free grasp planning in cluttered environments. The code and model are available at \href{https://github.com/robotic-manipulation/RNDF}{https://github.com/robotic-manipulation/RNDF}.
\end{abstract}
\section{Introduction}

Research indicates that humans rely on internal models~\cite{kawato1999internal, bryant1992internal} for interactive motion generation in the physical world. These models offer intuitive representations of spatial relationships between the body and the environment, enabling the simulation of potential movements and outcomes prior to taking physical action.

To enhance the efficiency of robot planning tasks, modeling spatial relations is crucial. However, this can be computationally intensive if approached analytically~\cite{gilbert1988fast}, particularly when dealing with highly irregular shapes of robot links. An alternative is to use simplified approximations, such as convex meshes~\cite{escande2007continuous} or shape primitives~\cite{sugiura2007real, zimmermann2022differentiable}, which offer a trade-off between geometric precision and computational efficiency.

Recent progress~\cite{liu2022regularized, chen2022fully, quintero2024stochastic, koptev2022neural, li2024representing} in data-driven robot modeling with the signed distance function (SDF) has demonstrated superiority in configuration-conditioned distance query efficiency and general applicability. Its differentiable nature grants direct access to derivatives in joint space for optimization-based downstream tasks. Recent applications have showcased its advantages in robot tasks such as human-robot interaction~\cite{liu2022regularized}, visual self-modeling~\cite{chen2022fully}, motion planning~\cite{quintero2024stochastic}, collision avoidance~\cite{koptev2022neural}, and dual-arm lifting~\cite{li2024representing}.

\begin{figure}
    \centering
    \includegraphics[width=\linewidth]{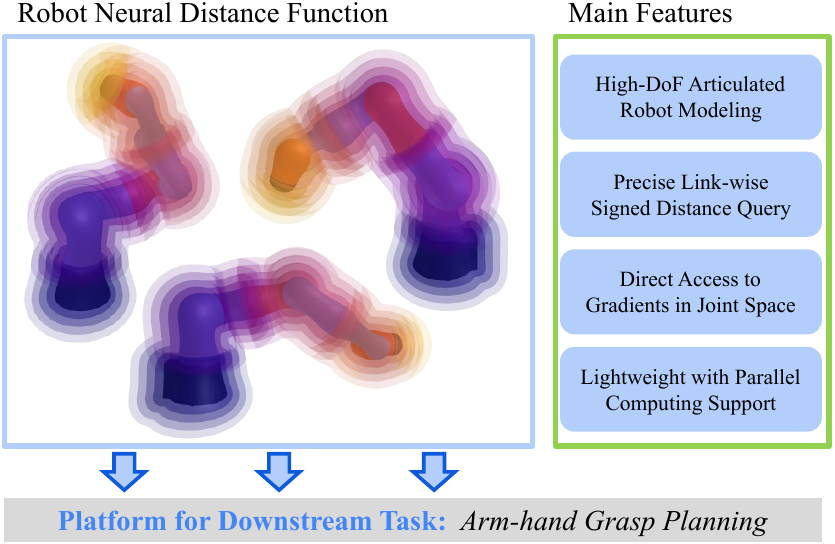}
    \caption{Visualization of the isosurface at different distance values of the proposed method, which showcases its precision and smoothness in 3D space. The inherent features of the proposed method benefit downstream tasks, and an arm-hand grasp planning framework is proposed.}
    \label{fig:fig1}
\vspace{-0.6cm}
\end{figure}

However, current SDF-based robot modeling methods still face the problem of balancing computational precision and efficiency due to the high dimensional nature of articulated kinematic chains. The geometry of the links that make up the robot and the angle of joints connecting these links determine the morphology of articulated robots. Following such a principle, current SDF-based robot modeling can be divided into two categories: 1. separate modeling of kinematic chains and link geometry and 2. joint modeling of robot geometry with forward kinematics.  \textbf{Separate modeling} represents the robot geometry, which combines independent SDF of each link based on the explicitly defined robot forward kinematics~\cite{li2024configuration, yang2024robotsdf, liu2023collision}. Though local SDF can model the link's geometry with high precision, explicitly following the forward kinematic chain step by step on each specific link requires a separate inference for the forward kinematic chain and the geometry of the link, which burdens the computational efficiency. By \textbf{joint modeling} robot kinematics and geometry with implicit neural function, the average computational time of each sample could be reduced from millisecond-level to microsecond-level due to parallel computing supported by the GPU~\cite{koptev2022neural, chase2020neural}. 
However, capturing the complex robot morphology in high-dimensional, non-linear configuration space poses a significant challenge. Since the configuration space is growing exponentially with the number of degrees of freedom, covering the target distribution with limited discrete sampled data is challenging. 
A significant limitation in previous works~\cite{koptev2022neural, quintero2024stochastic} using Multi-Layer Perceptrons (MLPs) is the substantial accumulation of SDF modeling errors along the forward kinematic chain. As reported in~\cite{koptev2022neural}, the error for the last link can be up to five times greater than that of the first link.

To address the aforementioned challenges, we propose the Robot Neural Distance Function (RNDF) (as shown in Fig.~\ref{fig:fig1}), which achieves millimeter-level configuration-conditioned morphology modeling on \textit{every} link for articulated robots with \textit{high} degree of freedoms (DoFs). Compared with the baseline method, RNDF achieves an \textbf{81.1\%} reduction of the average estimated distance error using only \textbf{47.6\%} model parameters. RNDF stands as a continuous distance representation within the robot configuration space and provides configuration-conditioned signed distance queries for each robot link for arbitrary points in the 3D space. The proposed method implicitly encodes robot geometry with forward kinematics, supports direct parallel computing, and remains fully differentiable. We evaluate the performance of RNDF with quantitative and qualitative comparisons in computational efficiency and accuracy with other configuration distance functions to highlight the advantages of our method.

To further showcase the capacity of the proposed method, we apply RNDF to an arm-hand system modeling for whole-body grasp planning. Different from most current research on robotic grasping solely focusing on finding grasp parameters first independently from the robot arm, we target grasp configuration planning on a unified arm-hand system~\cite{patel2023analysis, fang2024airexo}. An optimization-based framework is proposed by leveraging the 0-level set of RNDF for contact establishing and collision avoidance, which seamlessly integrates grasp synthesis, reachability check, collision avoidance, and motion planning for the holistic robotic arm-hand system.

In summary, the contribution can be summarised as follows:

\begin{itemize}
    \item We propose a high-precision, lightweight, and flexible configuration space signed distance function, Robot Neural Distance Function (RNDF), for jointly encoding robot geometry with forward kinematics. 
    \item We apply the proposed method to seven-DoF articulated robot modeling and highlight its advantages in both precision and efficiency with qualitative and quantitative evaluation.
    \item We further apply the proposed RNDF on arm-hand modeling and integrate it into optimization-based whole-body grasp planning within complex environments with obstacles.
\end{itemize}

\section{Related Work}
The Signed Distance Function (SDF)~\cite{osher1988fronts} is commonly used to quantify the nearest distance from a point of interest to the surface of a geometric shape. The sign of the distance delineates the point's position relative to the shape—either outside, on the boundary, or inside. Owing to its continuous and differentiable properties, the SDF offers a robust means to represent and manipulate geometry~\cite{park2019deepsdf}. This has found extensive applications in robotics, including robotic grasp synthesis~\cite{weng2023neural, liu2021synthesizing}, object shape modeling~\cite{chen2023sliding, gpis, egpis, stanimir}, and scene modeling~\cite{tekden2023neural, active_gpis}, among others~\cite{hu2021radiation}.
Commonly used approaches to construct SDFs include kernel methods and neural networks.

\textbf{Kernel methods} present a mathematically rigorous and adaptable technique for constructing SDFs~\cite{shawe2004kernel}. These methods represent a function as a weighted sum of basis functions (kernels) centered at specific data points. The choice of kernel function dictates the SDF's ability to represent various shape features. The efficacy of different kernels in shape modeling is assessed by~\cite{gandler2020object}. Additionally,~\cite{chen2023sliding} delves into the use of Radial Basis Functions (RBF) to steer tactile perception in unknown object shape modeling. Both~\cite{mahler2015gp} and~\cite{de2021simultaneous} harness the uncertainty inherent in the Gaussian process implicit surface (GPIS) for enhanced grasp planning. \cite{li2024representing} leverages Bernstein Polynomial (BP) for learning the geometry of each robot link and combines them with forward kinematics. \cite{maric2024online} propose an online learning scheme using piecewise polynomial basis functions.

\textbf{Neural networks}, in contrast to kernel methods, exhibit superior computational efficiency and adeptness at addressing intricate, highly non-linear challenges. Their integration with SDFs has ushered in a wide range of task-specific function learning. For instance, \cite{tekden2023neural} advocates for the fusion of SDF with motion primitives in robotic motion planning. \cite{driess2022learning} employs SDFs for robot manipulation planning, while \cite{ortiz2022isdf} pioneers online continuous learning of the SDF. By incorporating joint configuration as an input, SDFs can be extended to model intricate robot geometries. \cite{mu2021sdf} incorporates separate codes for encoding shape and articulation. \cite{koptev2022neural} introduces the concept of learning neural SDFs in robot joint space, facilitating distance-to-collision estimations for any given point and robot configuration. Similarly, \cite{liu2022regularized} leverages a joint space SDF for human-robot interaction, and \cite{gao2023realtime} employs multi-layer perceptrons to model robotic hand geometries for in-hand manipulation. 

The aforementioned methods demonstrate high accuracy in static and low-DoF object geometry modeling. However, as the dimensionality increases, modeling accuracy diminishes. While current implicit neural functions adeptly model objects or robotic fingers, their precision on robotic arms remains inadequate. Despite these methods capturing basic forward kinematics and fuzzy robot geometry, prediction errors escalate considerably along the forward kinematic chain. This limitation hinders contemporary techniques from accurately modeling high-DOF robotic systems.
In this paper, we address these challenges by incorporating the inductive biases of high-DOF manipulators into the network design, achieving refined robot geometry learning. The proposed method's high precision, lightweight, and support for direct parallel computing highlight its advantages over other modeling methods, unlocking potential applications in tasks requiring strict contact and non-contact constraints.

\section{Robot Neural Distance Function}

In this section, we outline the design of the proposed Robot Neural Distance Function, which implicitly encodes the robot geometry using forward kinematics in the configuration space.
\subsection{Problem Formulation}
We consider a serial links chain consisting of $m$ revolute joints and $n$ links. The robot configuration is defined by a combination of joint positions $\mathbf{q} = [q_1,...,q_m]$ in the configuration space $\mathcal{C}\in \mathbb{R}^m$. The robot's surface $\mathcal{S}_{R}(\mathbf{q})$ is conditioned on configuration $\mathbf{q}$, consisting of $n$ link surfaces $\mathcal{S}_{l, k}$, $k=1,2,\dots,n$.

Given an arbitrary point $\mathbf{p}=[x, y, z]\in \mathbb{R}^3$ in the 3D space, we denote the minimum distance from this point to the surface of the $k$-{th} robot link as $|d_k| \in \mathbb{R}^{+}_{0}$, $k=1,2,\dots,n$. The sign of each distance $d_k$ is determined as follows:
\begin{equation}
\text{sgn}(d_k)=\left\{
\begin{aligned}
& =+1,  & \text{if p is outside the k-th link} \\
& =0,  & \text{if p is on the k-th link} \\
& =-1, & \text{if p is inside the k-th link}
\end{aligned}
\right.
\end{equation}
Finally, we stack the point-link distances into a vector $\mathbf{d} = [d_1, ..., d_n]\in \mathbb{R}^n$. The objective signed distance function $f_{R}$ of the robot link-wise surface is defined as $f_{R}=(\mathbf{q}, \mathbf{p})=\mathbf{d}$.
Moreover, we use the Jacobian matrix $\mathbf{J}=\frac{\partial f_{R}(q,p)}{\partial q} \in \mathbb{R}^m \times \mathbb{R}^n$ to represent the gradient direction of distance in the configuration space, which guides the robot move towards or away from given point.

\subsection{Network Architecture}
\begin{figure*}
    \centering
    \includegraphics[width=0.85\textwidth]{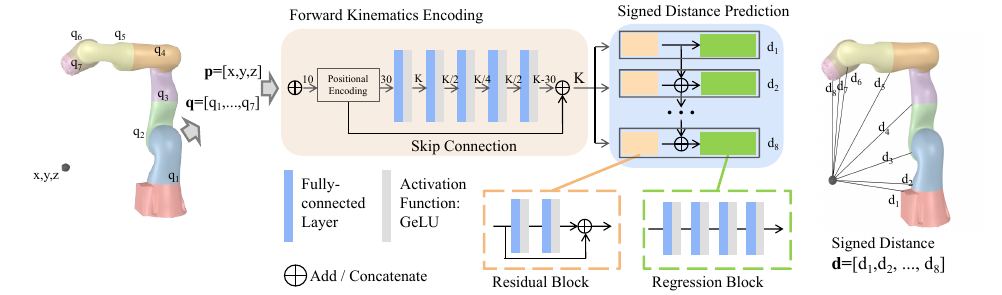}
    \caption{The detailed neural network structure of the proposed RNDF for KUKA iiwa 7, which constructs a mapping from $\mathbb{R}^{7+3} \rightarrow \mathbb{R}^{8}$. By incorporating the multi-head and hierarchical feature design, we achieve high-precision distance prediction with a significantly smaller model size.}
    \label{NN_struct}
    \vspace{-0.3cm}
\end{figure*}

\begin{figure}[h]
    \centering
    \includegraphics[width=0.8\linewidth]{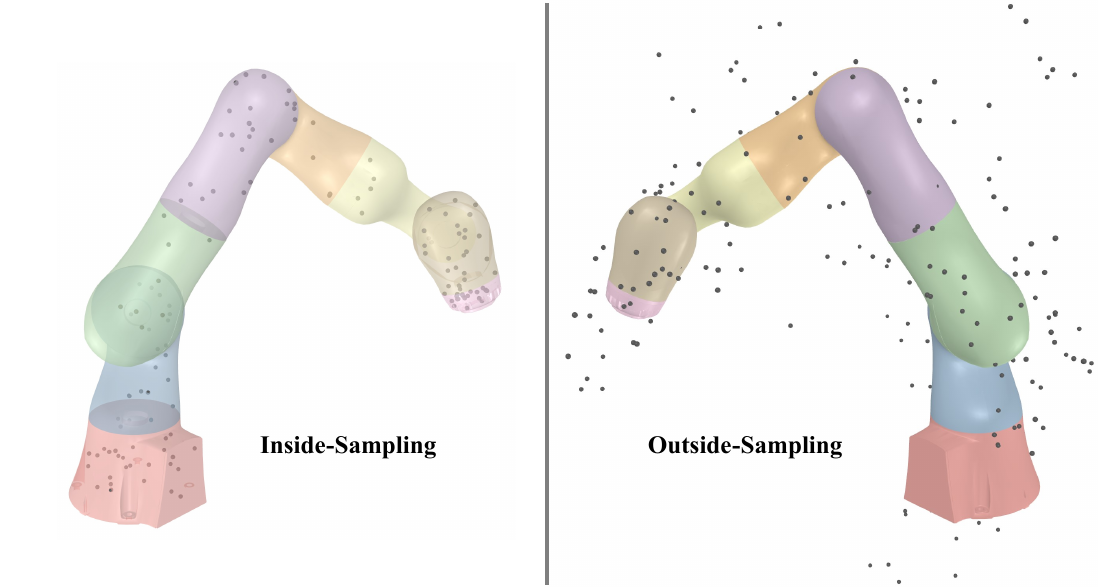}
    \caption{The data distribution is balanced through a separate inside and outside sampling design, ensuring accurate boundary modeling.}
    \label{fig:sample}
\vspace{-0.2cm}
\end{figure}

\begin{figure*}[h]
    \centering
    \includegraphics[width=0.9\textwidth]{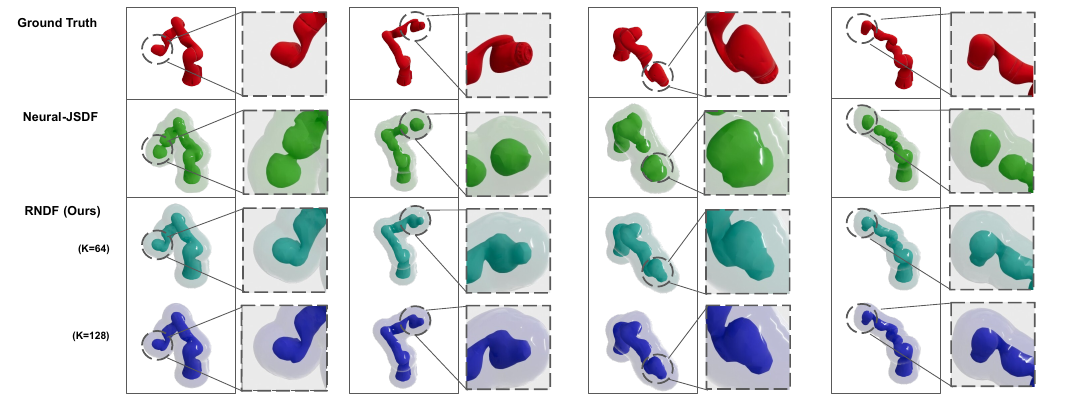}
    \caption{Examples of the visualized implicit distance function with random joint configurations built by different methods. The solid and transparent isosurfaces represent the prediction value $\text{min}(f(\mathbf{q},\mathbf{p}))=0.001$ and $\text{min}(f(\mathbf{q},\mathbf{p}))=0.1$ respectively. Our proposed RNDF (with different feature sizes) shows superiority in smoothness, stability, and precision.} 
    \label{fig:model_eval}
\vspace{-0.4cm}
\end{figure*}
The network is designed to approximate the target function $f_{R}(\mathbf{q}, \mathbf{p}) \rightarrow \mathbf{d}$, building a mapping $\mathbb{R}^{m+3} \rightarrow \mathbb{R}^n$. 
Previous works demonstrate that an MLP can capture the robot's coarse morphology. Despite all the fine details of robot link geometry vanishing as the sequence of robot links on the serial chain increases, MLP-based networks possess the ability to implicitly encode the forward kinematic chain with approximate robot geometry. Thus, deriving the fine-grained local geometry features from the training data becomes a coarse-to-fine refinement problem. Based on the mechanism of the articulated robot, rigid links do not encounter any deformation during relative motion that occurs along the serial chain. Each rigid link that consists of the robot should be regarded as an independent object for geometry modeling. Based on this principle, we propose adopting a \textit{multi-head structured} design with a matched number $n$ of regression heads as robot links. The k-th regression head is responsible for the prediction of the signed distance $d_k$ for the k-th robot link. Though each rigid link is regarded as an independent object, the pose of each rigid link in the 3D space is entirely determined by its previous link on the kinematic chain and its relative rotation angle. Inspired by this mechanism, we further hypothesize that the $(k+1)$-th regression head should be conditioned on the features from the $k$-th for signed distance $d_{k+1}$ prediction.
%

Premised on the above design, we present the network design of the proposed robot neural distance function. The large number of DoFs defines a configuration space that grows exponentially; therefore, learning a smooth distribution in the configuration space from limited sampled data will be crucial. An encoder-decoder structured MLP with skip-connection and position encoding is adopted for forward kinematics feature extraction, providing an approximated morphology refinement for the upcoming regression heads. Such an hourglass-shaped design can not only reduce the model complexity but also effectively avoid overfitting limited data to learn a smooth distribution. The shared forward kinematics feature is further passed to $n$ regression heads for each link's $d_k$ prediction. To integrate the design of feature transmission in the link sequence, the mid-level feature from each regression head is further concatenated to the adjacent next link along the kinematic chain. The structure for learning hierarchical features through each link aims to preserve fine-grained robot geometry for each link jointly with accurate robot kinematic configuration.

The detailed network architecture for the proposed RNDF is illustrated with an example of KUKA iiwa 7 in Fig.~\ref{NN_struct}. It is worth noting that our method could be easily applied to any articulate kinematic chain by matching the number of regression heads. After position encoding, the backbone network up-projects the $(m+3)$-dimensional input to a latent vector with dimension K as a global feature. Eight individual regression heads predict the signed distance for the robot's eight separate links. Each regression head consists of a residual block and a fully connected regression block. The residual block provides a mid-level feature for both the current and upcoming link from the chain, while each regression block only conditions on the mid-level features from itself and its previous link. We choose Gaussian Error Linear Units (GeLU) as an activation function to consider smoother decision boundaries. The root mean square error (RMSE) is adopted as the loss function.

\subsection{Signed Distance Data Generation}\label{subsec:dataset}

The sampled signed distance data is formulated based on the mapping relation $\mathbb{R}^{m+3} \rightarrow\mathbb{R}^n$. After sampling a point $\mathbf{p}\in\mathbb{R}^3$ given a robot configuration $\mathbf{q}\in\mathbb{R}^m$, the group of signed distances $\mathbf{d}$ from $\mathbf{p}$ to each link's surface is collected as training labels. Similar to \cite{koptev2022neural}, we perform uniform sampling within the joint limits of the robot to generate the training dataset with 5\% expansion on joint limits. To enhance accuracy near the robot's surface, we densely sample points within the distance threshold of $d_{s}$ near the robot's surface\footnote{The distance threshold could be adjusted accordingly for robots with a different scale.}. To avoid biasing the model, we balance the dataset by ensuring that half of the sampled points are inside the robot as illustrated in Fig.~\ref{fig:sample}.

\begin{table*}[h]
\centering
\caption{Performance Evaluation (\textbf{RMSE,mm}) comparing with baseline methods, \textbf{Close} query points with a distance $\leq$ 100 mm, and \textbf{far} ones with distance $>$ 100 mm.}
\label{tab:model_eval}
\resizebox{\textwidth}{!}{%
\begin{tabular}{|c|c|cc|cc|cc|cc|cc|cc|cc|cc|cc|c|}\hline
\multirow{2}{*}{Method} & \multirow{2}{*}{Num. of Params} & \multicolumn{2}{c|}{Link 1}    & \multicolumn{2}{c|}{Link 2}    & \multicolumn{2}{c|}{Link 3}    & \multicolumn{2}{c|}{Link 4}    & \multicolumn{2}{c|}{Link 5}    & \multicolumn{2}{c|}{Link 6}    & \multicolumn{2}{c|}{Link 7}    & \multicolumn{2}{c|}{Link 8}    & \multicolumn{2}{c|}{\textbf{Avg.}} & \multirow{2}{*}{\textbf{Avg. Acc.}}  \\ \cline{3-20}
                  &                & \multicolumn{1}{c|}{close} & far & \multicolumn{1}{c|}{close} & far & \multicolumn{1}{c|}{close} & far & \multicolumn{1}{c|}{close} & far & \multicolumn{1}{c|}{close} & far & \multicolumn{1}{c|}{close} & far & \multicolumn{1}{c|}{close} & far & \multicolumn{1}{c|}{close} & far & \multicolumn{1}{c|}{close} & far      &            \\ \hline Neural-JSDF \cite{koptev2022neural}
                  &          199,658         & \multicolumn{1}{c|}{6.4} & 8.1 & \multicolumn{1}{c|}{5.4} & 6.0 & \multicolumn{1}{c|}{6.5} & 7.7 & \multicolumn{1}{c|}{6.5} & 8.9 & \multicolumn{1}{c|}{7.1} & 10.5 & \multicolumn{1}{c|}{13.5} & 15.3 & \multicolumn{1}{c|}{9.6} & 14.1 & \multicolumn{1}{c|}{11.7} & 19.2 & \multicolumn{1}{c|}{8.4} & 11.2      & 0.93       \\ \hline Multi-head MLP
                  &        166,122         & \multicolumn{1}{c|}{3.1} & 4.9 & \multicolumn{1}{c|}{3.8} & 5.5 & \multicolumn{1}{c|}{4.0} & 6.1 & \multicolumn{1}{c|}{4.1} & 6.4 & \multicolumn{1}{c|}{4.4} & 7.0 & \multicolumn{1}{c|}{5.1} & 8.1 & \multicolumn{1}{c|}{6.1} & 10.1 & \multicolumn{1}{c|}{6.2} & 10.7 & \multicolumn{1}{c|}{4.6} & 7.4     & 0.96      \\ \hline  RNDF K=64 \textbf{(Ours)} 
                  &          \textbf{95,098}         & \multicolumn{1}{c|}{1.3} & 1.2 & \multicolumn{1}{c|}{1.2} & 1.2 & \multicolumn{1}{c|}{1.4} & 1.3 & \multicolumn{1}{c|}{1.4} & 1.5 & \multicolumn{1}{c|}{1.7} & 2.0 & \multicolumn{1}{c|}{2.2} & 2.1 & \multicolumn{1}{c|}{2.8} & 4.1 & \multicolumn{1}{c|}{1.7} & 2.3 & \multicolumn{1}{c|}{\textbf{1.7}} & \textbf{2.0}       & 0.98    \\ \hline  RNDF K=128 \textbf{(Ours)} 
                  &          243,466         & \multicolumn{1}{c|}{1.3} & 1.3 & \multicolumn{1}{c|}{1.3} & 1.3 & \multicolumn{1}{c|}{1.3} & 1.3 & \multicolumn{1}{c|}{1.3} & 1.3 & \multicolumn{1}{c|}{1.5} & 1.5 & \multicolumn{1}{c|}{1.7} & 1.6 & \multicolumn{1}{c|}{2.0} & 2.0 & \multicolumn{1}{c|}{1.5} & 1.7 & \multicolumn{1}{c|}{\textbf{1.5}} & \textbf{1.5}   & \textbf{0.99}  \\ \hline
\end{tabular}%
}
\end{table*}





\section{Numerical Evaluation}
We perform a comprehensive evaluation of multiple techniques for constructing the implicit distance function in joint space to emphasize the benefits of our design. The baseline method, Neural-JSDF, employs an MLP with position encoding. To delve deeper into the efficacy of our proposed multi-head structure and the hierarchical feature design, we introduce another neural network, termed \textit{Multi-head MLP}, as a variant of RNDF. While the multi-head MLP follows the same separate regression head design, the primary distinction between it and our RNDF lies in the former's absence of the mid-level features transmission design. All numerical experiments are carried out on a KUKA iiwa 7 robot arm with $\text{n}=8$ links with $\text{m}=7$ degrees of freedom.

\subsection{Qualitative Evaluation}
We sampled random configurations in the robot $\mathcal{C}$-space and visualized their corresponding isosurface with the prediction value $\text{min}(f(\mathbf{p}, \mathbf{q}))=0.001$ and $\text{min}(f(\mathbf{p}, \mathbf{q}))=0.1$. The result is illustrated in Fig.~\ref{fig:model_eval} with details amplification on the last link. The proposed RNDF (with feature sizes $\text{K}=64$ and $\text{K}=128$) is superior in smoothness and stability and can capture the fine-grained geometry of links with intricate shapes. The defect of error accumulation on the last link from previous methods has been addressed well, as the predicted isosurface of the last link is as precise as the first link on the kinematic chain. 

\subsection{Quantitative Analyze}
All testing data follows the identical distribution as the training data but is sampled independently with different random seeds. We sampled data with $\mathbf{p}=200,000$ and $\mathbf{q}=1,000$ for each evaluation with a specific distance requirement. An ablation study is performed on both prediction distance errors and boundary classification, investigating how the different design contributes to the model performance:

\paragraph{Distance Prediction Evaluation}
A detailed quantitative analysis of the model's performance on each link is listed in Table \ref{tab:model_eval}. Though we can observe a significant prediction error reduction compared with Neural-JSDF by integrating the design of multiple regression heads, the significant error accumulation from the first link to the last link still exists in the Multi-head MLP. After combining the design of separate regression heads with hierarchical feature transmission, the proposed RNDF, with a feature size of 64, achieves an error reduction of 81.12\% with the usage of only 47.6\% model parameters compared with Neural-JSDF. By increasing the dimension of the feature size $\text{K}$ up to 128, RNDF achieves a precision of $\textbf{1.5 mm}$ on a KUKA IIWA 7 robot with 800mm maximum reach length. 

\paragraph{Classification Accuracy Evaluation} 
To further delve into the performance within areas close to the robot's surface, we also evaluate the classification accuracy to distinguish between collided and free configurations with $d < 30$ mm (as shown in Table.~\ref{tab:model_eval}) Model performance on this property is crucial cause it directly determines its reliability on challenging collision avoidance and motion planning tasks. The average classification accuracy of $\text{sgn}(f(\mathbf{q}, \mathbf{p}))$ over all links is $\mathbf{99\%}$, showing evident improvement compared with Neural-JSDF \cite{koptev2022neural} and Multi-head MLP.  

\subsection{Computational Efficiency Evaluation}
The proposed RNDF can serve as a collision-checker in the $\mathcal{C}$-space for downstream applications such as sample-based path planning. We compare the computational performance of RNDF with the standard Gilbert-Johnson-Keerthi (GJK) algorithm \cite{gilbert1988fast}, which costs  0.6 $\mu$s to calculate one distance query between two convex hulls as reported in \cite{montaut2022collision}. Therefore, we use a baseline of 4.8 $\mu s$ for the 8-link robot arm distance-to-collision evaluation. The computational performance of the proposed RNDF with different feature sizes is listed in Table~\ref{compute_eval}. Our RNDF estimates signed distance for 8 individual links with a total time cost of only \textbf{0.183} $\mathbf{\mu s}$ by leveraging its differentiable nature (e.g. batch sampling in the $\mathcal{C}$-space or cartesian space), achieves a cost time reduction of \textbf{96\%} compared to standard GJK algorithm. All results are averaged on 1k runs with GeForce RTX 3070 GPU and an 16-core 2.30GHz CPU.
Experimental results show that our RNDF can serve as a solid platform for various applications.
\begin{table}[]
\centering
\caption{Computational Performance of RNDF}
\resizebox{\linewidth}{!}{%
\label{compute_eval}

\begin{tabular}{|c|cc|cc|}
\hline
\multirow{2}{*}{Method} & \multicolumn{2}{c|}{CPU}    & \multicolumn{2}{c|}{GPU}    \\ \cline{2-5} 
                  & \multicolumn{1}{c|}{batch time} & sample time & \multicolumn{1}{c|}{batch time} & sample time \\ \hline RNDF(k=64)
                  & \multicolumn{1}{c|}{0.303 $s$} & 3.031 $\mu s$ & \multicolumn{1}{c|}{18.319 $ms$} & \textbf{0.183} $\mu s$  \\ \hline RNDF(k=128)
                  & \multicolumn{1}{c|}{0.553 $s$} & 5.532 $\mu s$ & \multicolumn{1}{c|}{26.804 $ms$} & \textbf{0.268} $\mu s$ \\ \hline
\end{tabular}%
}
\begin{tablenotes}
\item[1] The batch size is set to 100,000 for testing. 
\end{tablenotes}
\vspace{-0.4cm}
\end{table}

\section{Applications to Arm-hand Grasp Planning}
In this section, we model an arm-hand system with the proposed method, consisting of a 7-DoF KUKA iiwa 7 robot arm $f_{arm}(\mathbf{q}_{arm}, \mathbf{p}^{arm})$ and a 16-DoF Allegro Hand $f_{hand}(\mathbf{q}_{hand}, \mathbf{p}^{hand})$. A novel model-based framework is proposed for diverse types of grasp planning while ensuring the output arm-hand grasp configuration is collision-free regarding its surrounding environments.
\subsection{Experimental Setup}
The system is modeled as six independent kinematic chains (one from the arm and five from the hand) because its high dimensionality can hinder the data sampling and training process. We view the robotic hand as a set of independent serial link chains as introduced in \cite{gao2023realtime}, separately train an RNDF for each kinematic chain and subsequently merge them \cite{Zhong_PyTorch_Kinematics_2024}. This separation also aids in addressing the scale imbalance issue.  

The hand's base frame is defined by arm forward kinematics and is represented using a transformation matrix $H \in SE(3)$.
To estimate the distance-to-collision for the hand links, points from the world frame are first transformed to the hand's base frame.

\subsection{Formulation for Grasp Planning}

\paragraph{Grasp Stability Constraints}

A widely used approach to assess grasp stability is based on the force-closure (FC) criteria \cite{nguyen1988constructing}. In this work, we adopt the point contact model, considering the force to be aligned with the object's surface normal at the contact. The contact point is approximated as the nearest point on the target object from the chosen link-in-contact. The stability constraint yields a binary result based on a pre-defined friction coefficient. Such constraint formulation could be easily combined with the concept of affordance area in robotic manipulation to generate task-relevant grasp configuration and confine the optimization process.

\paragraph{Collision-free Constraints}
This constraint directly estimates the distance to collision for the grasping task. Given the system's RNDF $f_{arm-hand}$ for links $\mathbf{L}_{arm-hand}$, and the selected link set $\mathbf{L}_c$ for contact establishing, we need to evaluate the collision constraints for two link groups individually: 
\begin{itemize}
    \item[1.] The minimum value of the predicted link-wise distances $\mathbf{d}_{arm-hand}^{obs}$ between the given obstacles $\mathbf{P}_{obs}$ and the system's links $\mathbf{L}_{arm-hand}$ should be larger than the given threshold $\mathbf{d}_{min}^{obs}$:
    \begin{equation}\label{eq:collision_free}
        \text{min}(\mathbf{d}_{arm-hand}^{obs}) \geq \mathbf{d}_{min}^{obs},
    \end{equation}
  to ensure a safe distance between the robotic system and its environment. 
    \item[2.] The free links set $\mathbf{L}_{f}= \mathbf{L}_{arm-hand} - \mathbf{L}_{c}$ that are not used to establish contacts for grasping w.r.t the target object $\mathbf{P}_{obj}$:
    \begin{equation}\label{eq:collision_free}
        \text{min}(\mathbf{d}_{f}^{obj}) \geq 0,
    \end{equation}
    where $\mathbf{d}_{f}^{obj}$ denotes the predicted distances between the free links and the target object. Its minimum value $\text{min}(\mathbf{d}_{f}^{obj})$ needs to remain positive to obstruct the possible penetration between the free links and the target object.
\end{itemize}

\paragraph{Objective Function}
The goal of grasp planning is to establish contact with the target object's surface under the above constraints. The objective function is designed to achieve this goal by minimizing the absolute value of the \emph{distance-to-collision} between the selected links $\mathbf{L}_{c}$ and the target object to establish contacts while keeping the system collision-free. A small penetration value $d_{p}$ is set to ensure secure contact with a limited force that can be applied from the selected links to the target object. The objective function is formulated as follows:
\begin{equation}
\mathcal{Q} = \lambda_1~\text{max}(\text{ReLU}(\mathbf{d}_{c}^{obj})) + \lambda_2~\text{max}(\text{ReLU}(-\mathbf{d}_{c}^{obj}-d_{p}))
\end{equation}
where $\lambda_1$ and $\lambda_2$ are the weights to balance the on-surface contact and the penetration depth $d_{p}$. $\mathbf{d}_{c}^{obj}$ denotes the predicted signed distances between the selected links $\mathbf{L}_{c}$ and the target object $\mathbf{P}_{obj}$.

\begin{table}[]
\centering
\caption{Arm-Hand Power Grasp Evaluation in Simulation}
\label{tab:power_test}
{
\begin{tabular}{|c|c|c|c|c|c|}
\hline
Evaluated Term & Power Grasp & Collision Avoidance \\ \hline
Successful Rate & 78\% & 100\% \\ \hline
\end{tabular}%
}

\end{table}

\subsection{Experimental Result}
We use SLSQP~\cite{kraft1988software} as our solver for objective function minimization. Before solving the optimization problem, we analyzed the hand kinematic model offline to construct the reachability map, collision map, and opposition links \cite{yao2023exploiting}. Based on the grasp taxonomy introduced by \cite{feix2015grasp}, we test the arm-hand grasp configuration planning ability on pinch and power grasp in cluttered environments (obstacles in surroundings). In addition, we also demonstrate the potential of arm-hand wrapping planning beyond using the end-effector but the whole upper limb for grasping.

\begin{figure}
    \centering
    \includegraphics[width=0.9\linewidth]{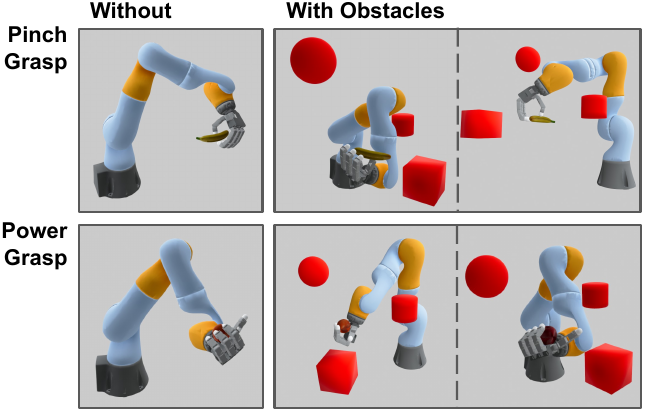}
    \caption{Examples of generated arm-hand pinch and power grasp configuration with different environmental settings. The red sphere, cube, and cylinder represent obstacles that the arm-hand system needs to avoid. The target object is a banana for pinch grasp and an apple for power grasp.}
    \label{fig:grasp}
\end{figure}

\begin{figure}
    \centering
    \includegraphics[width=0.9\linewidth]{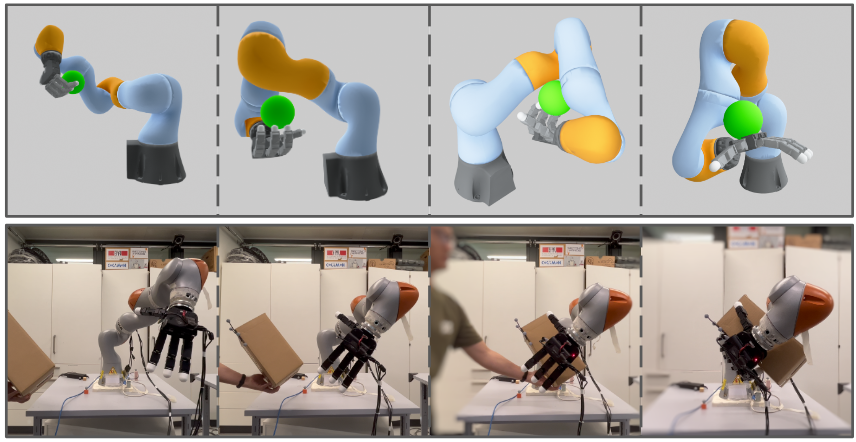}
    \caption{The top line of images shows the examples of the arm-hand wrapping a green ball. The bottom line of the images shows the handover of a large paper box from a human to a robot using arm-hand wrapping.}
    \label{fig:wrapping}
\vspace{-0.4cm}
\end{figure}

\paragraph{Pinch Grasp}
A pinch grasp is a type of grasp in which an object is held between the tips of the thumb and one or more fingers. This grasp is used for holding small objects with precision rather than force. The in-contact link is set to the fingertip link of the thumb and the index while ensuring collision-free constraints. The planning result is illustrated in the upper line of Fig.~\ref{fig:grasp}.
\paragraph{Power Grasp}
A power grasp refers to holding an object firmly in the palm and wrapping the fingers around it to exert significant force. The in-contact link is set to the entire links from the hand as constructing a finger-palm wrapping. The planning result is demonstrated in the bottom line of Fig.~\ref{fig:grasp}. We also evaluate the effectiveness of the planned arm-hand configuration within PyBullet. The experimental result of 50 trials is listed in Table.~\ref{tab:power_test}, which achieves a satisfactory successful gripping rate while ensuring applicable whole-body configurations.
\paragraph{Arm-hand Wrapping}
We also demonstrate the potential application of arm-hand wrapping based on the proposed framework, as shown in Fig.~\ref{fig:wrapping}. After the feasible opposition links are selected for grasping, our framework possesses the ability to perform grasp planning with the whole upper limb beyond the scope of the end-effector.

\section{Conclusion}

In conclusion, we introduce a novel robot morphology model, the Robot Neural Distance Function (RNDF), based on a signed distance function in configuration space, designed to enhance both speed and accuracy in planning tasks. Experimental results validate the effectiveness and precision of our approach. Additionally, we extend its application to a unified arm-hand configuration planning system and propose an advanced framework for high-DoF planning, incorporating force-closure and collision-free constraints.

We plan to extend the current work in the following directions: 
incorporating the dynamic model into arm-hand grasp planning tasks, employing a global optimization approach to avoid local minima from poor initialization, and refining the formulation to handle noisy point clouds effectively.

\newpage
\bibliographystyle{IEEEtran}
\bibliography{ref}

\end{document}